# Stro-VIGRU: Defining the Vision Recurrent-Based Baseline Model for Brain Stroke Classification

*Subhajeet Das, Pritam Paul, Rohit Bahadur, Sohan Das*
Dept. of CSE-AI, Brainware University, Kolkata, India

**Abstract**

Stroke majorly causes death and disability worldwide, and early recognition is one of the key elements of successful treatment of the same. It is common to diagnose strokes using CT scanning, which is fast and readily available, however, manual analysis may take time and may result in mistakes. In this work, a pre-trained Vision Transformer-based transfer learning framework is proposed for the early identification of brain stroke. A few of the encoder blocks of the ViT model are frozen, and the rest are allowed to be fine-tuned in order to learn brain stroke-specific features. The features that have been extracted are given as input to a single-layer Bi-GRU to perform classification. Class imbalance is handled by data augmentation. The model has achieved 94.06% accuracy in classifying brain stroke from the Stroke Dataset.

**Keywords**: Brain Stroke Classification, Vision Transformer, Bidirectional Gated Recurrent Unit, Transfer Learning, Medical Image Analysis

## I. Introduction

Brain stroke refers to an acute health problem exhibited by the disruption or slowdown of blood supply to the brain, causing the death of neurons and possibly incurable neurological injuries. It is still one of the major factors of worldwide death and permanent impairment. The World Health Organisation reports that every year, millions of people around the globe succumb to strokes, and timely screening is essential in increasing life-saving measures and reducing morbidity. Timely and correct diagnosis is a must in order to initiate prompt care that can entail thrombolysis therapy or surgery, both of which can help a great deal.

One of the common diagnostic tests that is mostly used in the process of detecting stroke is the Computed Tomography (CT) scan, as it gives a reliable result with its short time of acquisition. However, CT scans require manual interpretation by radiologists, making this process time-consuming, prone to error, and subjective because of the inherent complexities, especially in stroke cases that occur in the early stages, because the symptoms are not very pronounced. These issues echo the necessity of automated and intelligent diagnostic systems that could help medical personnel solve the problem of reliably and promptly classifying a stroke condition.

This article presents a new hybrid deep learning algorithm that uses the strengths of Vision Transformer (ViT) [1] and Bidirectional Gated Recurrent Units (Bi-GRU) to recognize brain strokes using CT scan images. The pre-trained ViT model is employed to pull out relevant spatial features from CT-scan samples comprehensively. The extracted embeddings are thereby fed to the single-layered Bi-GRU network that captures the temporal and sequence relationship between the visual features, leading to good classification performance. We use different data augmentation techniques on the training set to resolve the issue of class imbalance in the dataset. With this setup, the transfer learning model has achieved 94.06% accuracy.

## II. Literature Review

Artificial intelligence (AI) has become a key tool in brain stroke detection improvement, mainly by using deep learning techniques in medical imaging [2]. Çınar et al. [3] explored the use of transfer learning with pre-trained ResNet101 [4], VGG19 [5], EfficientNet-B0 [6], MobileNet-V2 [7], and GoogleNet [8] models for the detection of brain strokes from CT scans. The study determined the relative performance of different architectures and suggested potential models for stroke identification.

Çınar et al. [9] also suggested a hybrid system involving deep feature extraction with CNNs and traditional machine learning algorithms, such as the support vector machine algorithm and K-nearest neighbors algorithm, thus suggesting improved performance in discriminating between ischemic, haemorrhagic, and normal brain images.

Uzun et al. [10] proposed a real-time graphical user interface (GUI) system with deep learning for the automatic segmentation and identification of strokes from CT imaging. The focus of their work was applying YOLO-based [11] real-time object detection networks with the advantages of utilizing more recent segmentation variants to improve detection efficiency and clinical use.

Katar et al. [12] proposed a Vision Transformer (ViT) model [1] for localisation and classification of stroke. Their work demonstrated the capability of transformer-based models and visualisation methods in clinical decision support and interpretability.

### III. Methodology

This research work proposes the Stro-VIGRU model, a hybrid transfer learning framework that combines the pre-trained Vision Transformer (ViT) [1] with a Bidirectional Gated Recurrent Unit (BiGRU) for visual sequence modelling for brain stroke classification. The selection of Bi-GRU in sequential feature modelling of the proposed Stro-VIGRU model is boosted by the fact that they provide similar learning capacity as LSTMs, but they are computationally faster as they use fewer gates compared to LSTMs, which means that they have fewer parameters to optimize and train faster, which is critical in the context of high-dimensional patch embeddings obtained with a Vision Transformer. Also, the bidirectional approach allows the model to incorporate the dependency in both forward and backward directions, which brings contextual interpretation of the spatial sequences. The complete schematic diagram of the Stro-VIGRU model is shown in Fig. 1.

Given an input image $I \in \mathbb{R}^{H \times W \times C}$, the ViT backbone divides it into a sequence of patches as described in Eqn. 1.

$$X = \{x_1, x_2, \ldots, x_P\}, \quad x_i \in \mathbb{R}^{P_s \times P_s \times C} \quad (1)$$

Where each patch is linearly projected into an embedding.

$$z_i^0 = E\, x_i + E_{pos}, \quad z^0 \in \mathbb{R}^{P \times D_{vit}} \quad (2)$$

Where, in Eqn. 2. $E$ is the patch embedding projection and $E_{pos}$ denotes positional encodings. The classification token is discarded as the ViT model is only used for extracting features from the images, effectively resulting in:

$$Z_{ViT} = ViT - Encoder(z^0) = \{z_1, z_2, \ldots, z_P\},$$
$$Z_{ViT} \in \mathbb{R}^{P \times D_{vit}} \quad (3)$$

To preserve robust low-level features while limiting training cost, the patch embedding layer and the initial $N = 6$ encoder blocks are frozen, as denoted by Eqn. 4.

$$\theta_{frozen} = E, Encoder_1, \ldots, Encoder_N \quad (4)$$

A learnable linear projection layer bridges the ViT output to the Bi-GRU input dimensions as shown in Eqn. 5.

$$Z_{bridge} = W_{bridge} Z_{ViT} + b_{bridge} \quad (5)$$

where:

$$W_{bridge} \in \mathbb{R}^{D_{gru} \times D_{vit}},$$
$$b_{bridge} \in \mathbb{R}^{D_{gru}}.$$

Here, $D_{vit} = 768$ and $D_{gru} = 512$. The sequence $Z_{bridge}$ is processed by a Bidirectional GRU (Bi-GRU) to capture contextual dependencies as described by Eqn. 6.

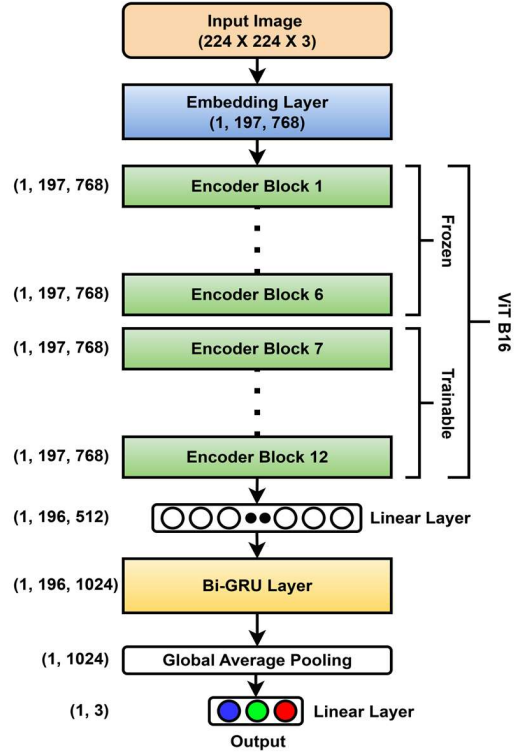

*Fig 1. Architecture of the Stro-VIGRU Model.*

$$H = \{h_1, h_2, \ldots, h_P\}, \quad h_i = [\vec{h}_i; \overleftarrow{h}_i] \quad (6)$$

with:

$$\vec{h}_i = GRU_{fwd}(z_{bridge,i}),$$
$$\overleftarrow{h}_i = GRU_{bwd}(z_{bridge,i}).$$

The output $H$ has dimension $H \in \mathbb{R}^{P \times 2D_{gru}} = \mathbb{R}^{P \times 1024}$. A temporal mean pooling is applied to obtain a fixed-length representation:

$$h_{pool} = \frac{1}{P}\sum_{i=1}^{P} h_i, h_{pool} \in \mathbb{R}^{2D_{gru}}, 2D_{gru} = 1024 \quad (7)$$

The final logits are computed by:

$$\hat{y} = W_{cls} h_{pool} + b_{cls}, \hat{y} \in \mathbb{R}^C, \quad (8)$$

where *C = 3* represents the number of classes in the brain stroke dataset.

The Stro-VIGRU framework thus combines powerful spatial representation learning of pre-trained ViT with sequential dependency modelling of Bi-GRU for providing a robust end-to-end solution for visual sequence analysis in brain stroke classification.

## IV. Experimental Setup

### A. Dataset Used

For training and evaluating the effectiveness of the proposed method, the Stroke Dataset [13] has been used. The dataset comprises of 6,650 images in total, distributed in three classes, namely the Bleeding, Ischemia, and Normal. Ischemia refers to the condition when there is a blockage in blood vessels, supplying oxygenated blood into the brain, and bleeding refers to the rupture of blood vessels, causing haemorrhage. The data is divided into a non-overlapping train and test set in an 8:2 ratio.

### B. Data Preprocessing

The image samples of the dataset are first reshaped to 224 × 224 × 3 pixels, and they are also normalized to have the pixel intensity within the range [0, 1].

### C. Data Augmentation

The Stroke dataset [13] contains three classes, where the 'Bleeding' class has 1,093 images, the 'Ischemia' class has 1,130 images, and the 'Normal' class has 4,427 images. To deal with this data imbalance problem, the method utilizes horizontal, vertical flipping, and 90° rotations with 50% probability, change in brightness and contrast with 30% probability, and median blur with 40% probability. These augmentation techniques are applied batch-wise to reduce resource requirements.

### D. Hyperparameters

The Stro-VIGRU model is trained for *200* epochs in total with a batch size of *32* and $1e^{-3}$ learning rate, as shown in Table 1. The Cosine Annealing scheduling algorithm gradually decreases the learning rate based on the cosine curve throughout the training process. Loss is calculated using the categorical cross-entropy loss, and the parameter-wise learning rate of the model is optimized using the Adam optimizer.

**Table 1:** Hyperparameters of the Model

| Hyperparameters | Values |
|---|---|
| Batch Size | 32 |
| Learning Rate | 1e-3 |
| Epoch | 200 |
| Optimizer | Adam |
| Scheduling Algorithm | Cosine Annealing |
| Minimum Learning Rate for Scheduler | 1e-6 |
| Loss Function | Categorical Cross-Entropy |

## V. Result and Discussion

While the proposed Stro-VIGRU model was evaluated against the non-overlapping test set of the Stroke Dataset [13], it achieved 94.06% of Top-1 accuracy, 98.80% of Top-2 accuracy, 93.99% of precision, 94.06% of Recall, and 93.97% of F1-Score, respectively. The class-wise performance of the model can be observed from Table 2. The epoch-wise train and test set accuracy curve is shown in Fig. 2.

**Table 2:** Results and Observations of the Model

| Class | Precision(%) | Recall(%) | F1-Score(%) |
|---|---|---|---|
| Bleeding | 93 | 93 | 93 |
| Ischemia | 91 | 82 | 86 |
| Normal | 95 | 97 | 96 |

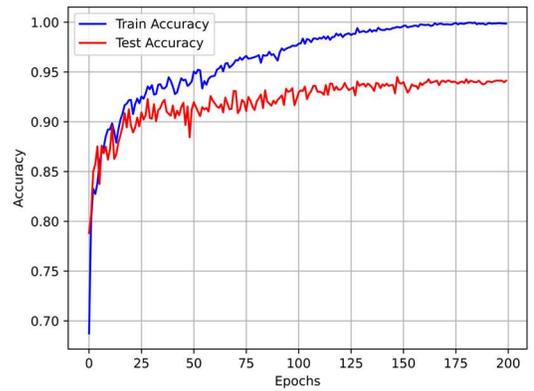

*Fig 2. Training vs. Test of the Stro-VIGRU Model.*

It is evident from the table that the method is struggling to correctly identify the Ischemia class because of high inter-class similarity with the Bleeding class, which can be observed from the UMAP [14] plot of the model, where the clusters of both classes are overlapping, as shown in Fig. 3.

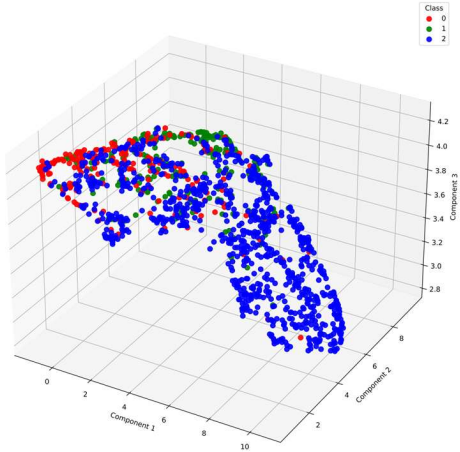

*Fig 3. UMAP plot of the Stro-VIGRU Model.*

## V. Conclusions

In this article, the Stro-VIGRU model is proposed, which combines a pre-trained Vision Transformer (ViT) and a Bidirectional Gated Recurrent Unit (Bi-GRU) to formulate a hybrid transfer learning method for early identification of brain stroke by using CT-scan data. The proposed model utilises the pre-trained ViT model to extract relevant features from the Stroke Dataset by freezing some layers to transfer the knowledge learnt from the ImageNet dataset, while the rest of the layers are made trainable to allow them to adapt to this dataset for effective feature extraction. The final trained model has achieved 94.06% accuracy, and it can be used for clinical applications. The study will be useful in the development of intelligent and automated systems for detecting strokes that could serve radiologists to make faster and more informed diagnoses, leading to improved care and patient diagnosis in time-sensitive medical practices.

The proposed Stro-VIGRU model showcases good performance in the context of brain stroke detection using CT-scan data. However, it has certain limitations. The imbalance in the dataset and inter-class similarity, especially between ischemia and bleeding, and the consideration of only CT scan images may have influenced the performance of the model.

The future improvements will include the addition of multimodal imaging data with CT scans, such as MRI, that can enhance the accuracy of the stroke diagnosis. A large dataset might enrich the variety of stroke outcomes, the demographics of the patients, and the conditions encountered by a patient during real-world clinical practice can be used, so that the model can become more generalised and robust.